\documentclass[10pt,twocolumn,letterpaper]{article}

\usepackage[pagenumbers]{wacv} 
\usepackage[accsupp]{axessibility}

\usepackage{multirow}
\usepackage{algorithm}
\usepackage{algorithmic}
\usepackage{multirow}
\usepackage{booktabs}
\usepackage{verbatim}
\usepackage{natbib}
\usepackage{tabularx}
\usepackage{booktabs}
\usepackage {tcolorbox}

\definecolor{wacvblue}{rgb}{0.21,0.49,0.74}
\usepackage[pagebackref,breaklinks,colorlinks,allcolors=wacvblue]{hyperref}

\def\wacvPaperID{682} 
\def\confName{WACV}
\def\confYear{2026}

\title{TA-Prompting: Enhancing Video Large Language Models \\for Dense Video Captioning via Temporal Anchors}

\author{
    {Wei-Yuan Cheng}$^{*1, \dagger}$, 
    {Kai-Po Chang}$^{*1}$, 
    {Chi-Pin Huang}$^1$, 
    {Fu-En Yang}$^2$, and {Yu-Chiang Frank Wang}$^{1,2,\ddagger}$\\
    \normalsize \textsuperscript{1} Graduate Institute of Communication Engineering, National Taiwan University\\
    \normalsize \textsuperscript{2} NVIDIA\\
    {\tt\small $^{\dagger}$r12942104@ntu.edu.tw, $^{\ddagger}$frankwang@nvidia.com} \\
}

\begin{document}
\maketitle

\def\thefootnote{*}\footnotetext{Equal contribution.}

\begin{abstract}

Dense video captioning aims to interpret and describe all temporally localized events throughout an input video. Recent state-of-the-art methods leverage large language models (LLMs) to provide detailed moment descriptions for video data. However, existing VideoLLMs remain challenging in identifying precise event boundaries in untrimmed videos, causing the generated captions to be not properly grounded. In this paper, we propose TA-Prompting, which enhances VideoLLMs via Temporal Anchors that learn to precisely localize events and prompt the VideoLLMs to perform temporal-aware video event understanding. During inference, in order to properly determine the output caption sequence from an arbitrary number of events presented within a video, we introduce an event coherent sampling strategy to select event captions with sufficient coherence across temporal events and cross-modal similarity with the given video. Through extensive experiments on benchmark datasets, we show that our TA-Prompting is favorable against state-of-the-art VideoLLMs, yielding superior performance on dense video captioning and temporal understanding tasks including moment retrieval and temporalQA.

\end{abstract}
    
\section{Introduction}
\label{sec:intro}

\begin{figure}[t]
  \centering
   \includegraphics[width=1\linewidth]{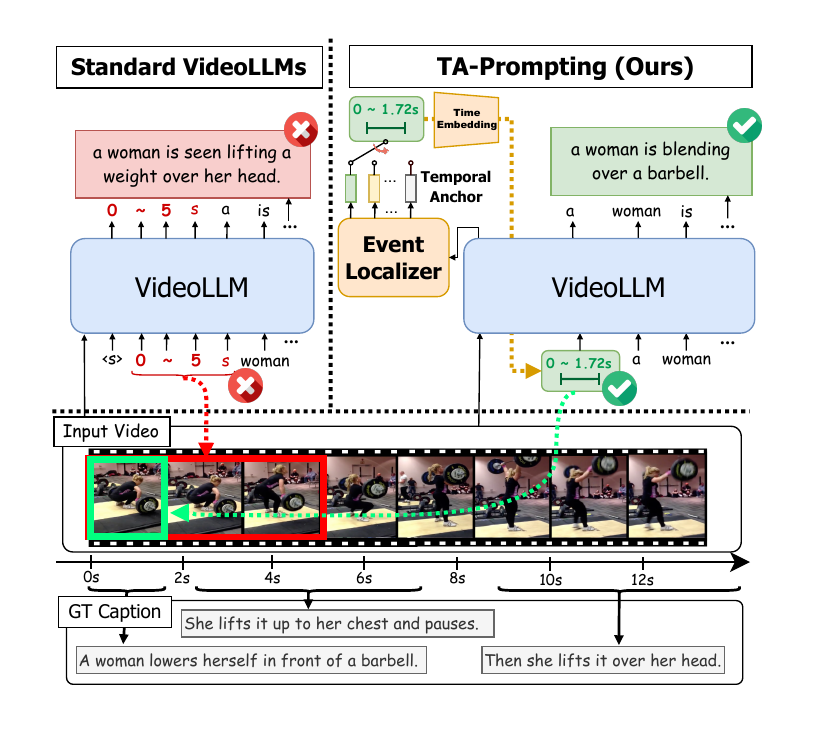}
   \caption{\textbf{Comparisons between standard VideoLLMs and \textit{TA-Prompting}.} Instead of using text tokens to describe time information, our \textit{TA-Prompting} employs temporal anchors to precisely localize events, which steers VideoLLM to caption the grounded video segment.}
   \label{fig:teaser}
\end{figure}

Videos contain abundant and layered information across various levels of granularities, from atomic actions to sequences of activities and overarching long-term goals. For example, in a video of a basketball game, the video includes the individual actions of players, like dribbling or shooting, and the intense process between teams as they attack and defend. A comprehensive caption for such a video should include both the granular details and the broader context of the game flow. However, existing video captioning models~\cite{fu2021violet, li2023lavender, wang2023all} tend to focus on individual moments within the video, often generating captions like ``a basketball player dribbles" without recognizing or articulating other concurrent and sequential events. To tackle this issue, dense video captioning is proposed~\cite{krishna2017dense, chen2021towards}, which aims to temporally localize multiple events and describe them throughout a video. By encouraging models to detect and describe multiple events, dense video captioning provides a more comprehensive and temporally fine-grained description of video content. As a result, dense video captioning has emerging a promising task within the vision-language learning field~\cite{wang2021end, suin2020efficient, wang2023learning}.

Early works~\cite{wang2021end, chen2021towards, wang2023learning} for dense video captioning primarily focus on identifying and describing individual events within a video, often treating each event as an independent instance. One of the most prominent approaches, PDVC~\cite{wang2021end}, employs a query-based transformer~\cite{carion2020end} to extract event features from video content. These event features are then used to predict both the event captions and timestamps. Building upon PDVC~\cite{wang2021end},~\citet{wang2023learning} further aligns the event features with their corresponding captions to encourage the output captions that are more contextually grounded within their respective events. In addition, DIBS~\cite{wu2024dibs} is proposed to enhance PDVC~\cite{wang2021end} by leveraging unlabeled video data in a semi-supervised learning manner.
Despite the advancements achieved by these methods, their generated captions for different events often lack narrative coherence. This issue arises because each caption is generated independently from individual event features, which restricts the capability of the model to capture the sequential dependencies between events~\cite{mun2019streamlined, wang2020event}.

To address the challenge of narrative inconsistency across event captions, recent approaches~\cite{li2023videochat,maaz2023video,huang2024vtimellm,ren2024timechat,qian2024momentor,huang2024lita,guo2024vtg, chang2025mitigating} have leveraged Large Language Models (LLMs) to enhance coherence in event captions by capitalizing on the autoregressive decoding capability of LLMs. Pioneering methods such as VideoChat~\cite{li2023videochat} and VideoChatGPT~\cite{maaz2023video} empower ImageLLMs~\cite{liu2024visual} through training them on introduced video-instruction tuning datasets. While these VideoLLMs are able to follow the language instructions, they remain struggling to accurately detect event boundaries due to the lack of temporally aware data within their instruction tuning datasets. 
To improve the temporal sensitivity of VideoLLMs, VTimeLLM~\cite{huang2024vtimellm} and TimeChat~\cite{ren2024timechat} augment existing video-instruction tuning datasets with timestamp annotations, so that the LLMs are tuned to directly generate event captions with their corresponding time spans. However, as recent works~\cite{chen2024rextime,leng2024mitigating,chen2024halc,wang2024mitigating} unveiled, the outputs of LLMs are prone to relying on language prior instead of focusing on the input video data, resulting in unfaithful event descriptions due to inaccurate timestamps. Hence, steering VideoLLMs to accurately localize temporal events for reasoning captions coherently over the whole video remains a challenging yet unsolved problem.

In this paper, we propose \textit{TA-Prompting}, which enhances  \textit{Video Large Language Models via \textbf{T}emporal-\textbf{A}nchors } for dense video captioning. Our \textit{TA-Prompting} employs a two-stage training scheme that first localizes all possible events within a video and then steers VideoLLMs to generate corresponding event captions. More specifically, \textit{TA-Prompting} involves an event localizer to predict a set of \emph{temporal anchors}, which represent the temporal spans of the identified events. Once these temporal anchors are obtained, we are capable of captioning each localized event by the conditions of temporal anchors. Furthermore, with the aim of enabling temporal coherence across all event captions, we employ a novel \textit{Event Coherent Sampling} strategy by selecting event captions with the highest confidence during inference time. As a result, globally coherent and locally dense event descriptions could be generated with precisely localized timestamps.

We now summarize the contributions of this work below:
\begin{itemize}
    \item We present \textit{TA-Prompting}, which enables event-based video captioning and understanding by localizing precise event timestamps while generating dense captions grounded to the segmented video clips.
    \item Our \textit{TA-Prompting} employs an event localizer to predict temporal anchors, which not only enables precise event localization but also steers VideoLLMs to generate temporal-aware dense video captions.
    \item Instead of applying traditional cross-entropy loss, we uniquely learn our event localizer by designing an anchor loss together with a temporal denoising objective, which significantly enhances the temporally grounded ability.
    \item We introduce event coherent sampling to tackle an arbitrary number of events during inference time and enhance the temporal coherence across event captions.
\end{itemize}
\section{Related Works}
\label{sec:related_work}

\begin{figure*}[t]
    \centering
    \includegraphics[width=0.95\linewidth]{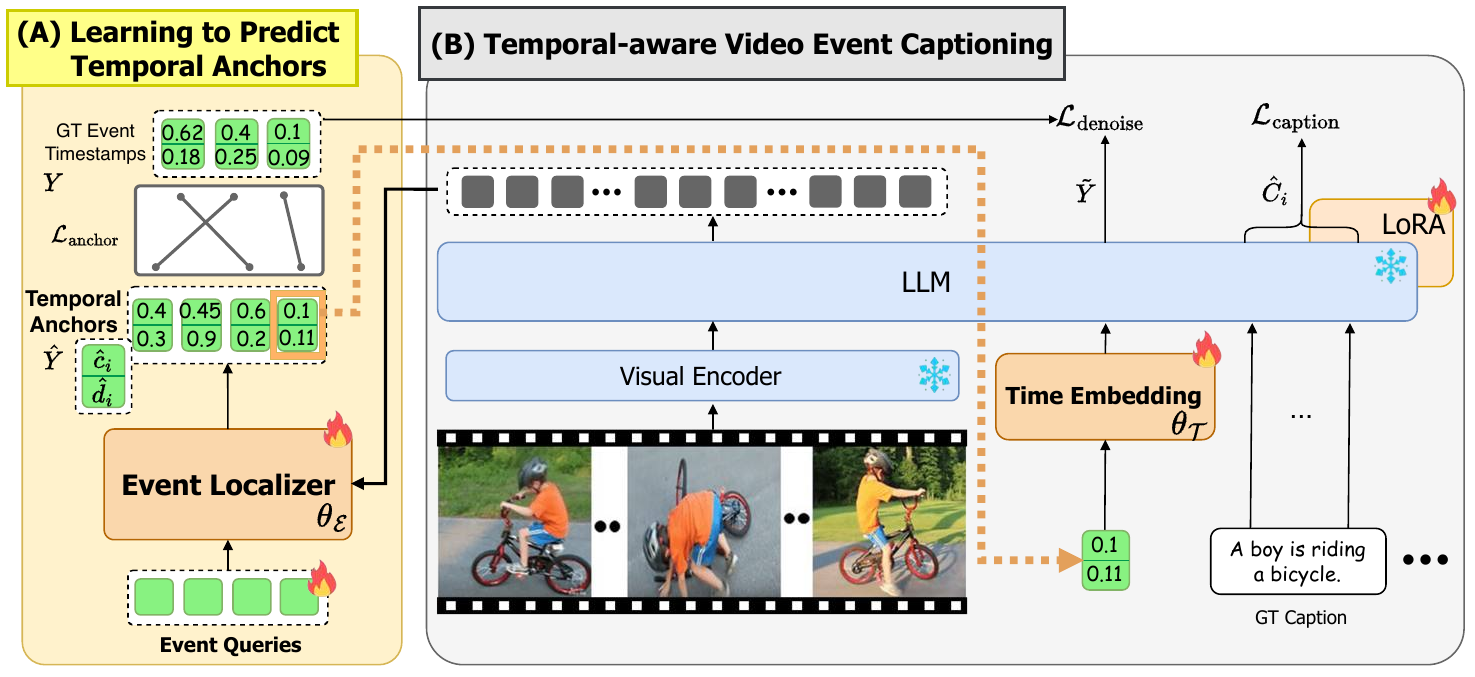}
    \caption{\textbf{Overview of \textit{TA-Prompting}.} 
    (A) \textit{TA-Prompting} learns to predict temporal anchors $\hat{Y}$ for each event in a video with an event localizer $\theta_{\mathcal{E}}$. (B) With temporal anchors as proposals, \textit{TA-Prompting} performs precise event localization and predicts temporal-aware event captions. Note that a time embedding layer $\theta_{\mathcal{T}}$ and a LoRA are jointly trained with $\theta_{\mathcal{E}}$ in (B).
    }
    \label{fig:main}
\end{figure*}
\vspace{-3mm}

\subsection{Video Large Language Model}

Recently, large vision language models~\cite{li2023blip, xu2024pllava, zhang2024video, wang2024internvideo2, wang2024qwen2, liu2024nvila, chang2024rapper} have evolved by leveraging the comprehensive understanding and reasoning capabilities of large language models (LLMs)~\cite{dubey2024llama, yang2024qwen2} to perform multimodal tasks. Building upon these successes, Video large language models (VideoLLMs), such as Video-LLaMA~\cite{cheng2024videollama}, Video-LLaVA~\cite{lin2023video} and LLaVA-Next-Video~\cite{zhang2024llavanextvideo}, have emerged from post-training image-based LVLMs on video-caption instruction tuning data. However, due to the limited context length of LLMs, the above VideoLLMs can only process short clips without sufficient capability to handle long videos, which contain multiple events. To overcome these limitations, several works~\cite{jin2024video, li2025llama, zhang2024long, xue2024longvila, shen2024longvu} advance video representation learning to efficiently represent a long video. For example, Video-LaVIT~\cite{jin2024video} decomposes a video into keyframes and motion vectors to represent spatial and temporal information, respectively. Nevertheless, the aforementioned methods tend to give a rough description of the entire video and remain challenging in detecting specific events that occur in a video along with their associated accurate timestamps.

\subsection{Event-Based Captioning and Understanding}

Event-based video understanding aims to precisely localize the temporal events while providing detailed captions for each relevant interval, which involves tasks of dense video captioning~\cite{caba2015activitynet, zhou2018towards}, moment retrieval~\cite{lei2021detecting, jang2023knowing, moon2023query, lin2023univtg} and TemporalQA~\cite{huang2024lita}. Several recent works~\cite{huang2024vtimellm,ren2024timechat,huang2024lita,guo2024vtg,qian2024momentor} have explored improving the temporal awareness of VideoLLMs. For example, VTimeLLM~\cite{huang2024vtimellm} and TimeChat~\cite{ren2024timechat} directly utilize textual tokens to represent time and train models with time-aware video-caption instruction tuning data. In addition, LITA~\cite{huang2024lita} and VTG-LLM~\cite{guo2024vtg} introduce learnable time embeddings to represent temporal information. Building on this, Momentor~\cite{qian2024momentor} further adopts a temporal-aware training strategy to ensure the representations of adjacent timestamps are similar. Despite improving the temporal awareness of VideoLLMs, the outputs of LLMs still tend to rely on language prior instead of grounding on video inputs, causing inaccurate and unfaithful output predictions. To tackle this challenge, we propose a unique \emph{Temporal-Anchored Video Large Language Models} to reason events within a video via the proposed temporal anchors and then steer VideoLLMs to generate temporally coherent captions grounded to the localized video segments.
\section{Method}
\label{sec:method}

\subsection{Problem Formulation}

Given an input video $v$, a dense video captioning dataset contains a set of event timestamp-caption pairs, denoted as $\{ (t_i^s, t_i^e, C_i) \}_{i=1}^{N}$, where \( t_i^s, t_i^e, C_i \) specify the starting time, ending time, and the corresponding caption of an event. $N$ is the number of events that occurred in the video $v$. The goal of the dense video captioning model is to predict the multiple event timestamps with their corresponding captions. As depicted in Fig.~\ref{fig:main}, we propose \textit{TA-Prompting} which exploits temporal anchors from the input video for improved dense video captioning. To boost the inference efficiency, we further propose an \textit{Event Coherent Sampling} scheme, which advances temporal coherence across video segments for caption prediction.

\subsection{Temporal-Anchored VideoLLMs}
\label{method:train}

\subsubsection{Learning to Predict Temporal Anchors}

\label{train:stage1}

With the goal of localizing all possible events within a video, we introduce an event localizer to predict temporal anchors $\hat{Y}=\{(\hat{c}_i, \hat{d}_i)\}_{i=1}^M$, where $(\hat{c}_i, \hat{d}_i)$ is the predicted time center and duration (span) for the event $i$, both value normalized to $[0,1]$ with respect to the video length. Inspired by DETR~\cite{carion2020end}, our event localizer is learned to output $M$ temporal anchors $\hat{Y}$ by employing learnable event queries to capture temporal boundaries from the given video. In practice, the event localizer is implemented as a lightweight transformer decoder with $M$ learnable event queries. Each query interacts with the video features through a cross-attention mechanism and outputs a predicted temporal anchor $(c_i, d_i)$.

As illustrated in Fig.~\ref{fig:main}, the event localizer, parameterized as $\theta_{\mathcal{E}}$, is trained to minimize the proposed anchor loss $\mathcal{L}_{\text{anchor}}$: 

\begin{align} \label{eq:loss1}
    \mathcal{L}_{\text{\scriptsize anchor}}(\hat{Y},Y; \theta_{\mathcal{E}}) &= \lambda_{\text{\scriptsize 1}} \mathcal{L}_{\text{\scriptsize seg}} + \lambda_{\text{\scriptsize 2}} \mathcal{L}_{\text{\scriptsize span}}.
\end{align}
In Eq.~\eqref{eq:loss1}, the segment loss $\mathcal{L}_{\text{seg}}$ is defined as:
\begin{align} 
    \mathcal{L}_{\text{\scriptsize seg}} = \sum^{N}_{i=1} \left( | \hat{c}_i-c_i |  + | \hat{d}_i-d_i | \right),
\end{align}
Note that $Y=\{(c_i, d_i)\}_{i=1}^N$ denotes the ground truth time center and span for each event, which can be simply derived from its start and end times $(t_i^s, t_i^e)$. To perform matching between $M$ time anchor predictions and $N$ ground-truth ones (we set $M$ $>$ $N$), the Hungarian Matching algorithm~\cite{kuhn1955hungarian} is utilized to pair the matched pairs by estimating the minimal $\mathcal{L}_{\text{anchor}}$.


On the other hand, the span loss $\mathcal{L}_{\text{span}}$ in Eq.~\eqref{eq:loss1} is to encourage the matching between the length of each predicted anchor $\hat{Y}$ and that of the ground-truth ones. We advance the concept of IoU~\cite{zheng2020distance} to define the span loss $\mathcal{L}_{\text{span}}$ as:
\begin{equation} \label{eq:gIoU}
\mathcal{L}_{\text{span}} = \sum^{N}_{i=1} \left[ 1 - 
\frac{
    \operatorname{max} \left( 0, \operatorname{min}(\hat{t}_{i}^{\text{e}}, t_{i}^{\text{e}}) 
    - \operatorname{max}(\hat{t}_{i}^{\text{s}}, t_{i}^{\text{s}}) \right)
}{
    \operatorname{max}(\hat{t}_{i}^{\text{e}}, t_{i}^{\text{e}}) 
    - \operatorname{min}(\hat{t}_{i}^{\text{s}}, t_{i}^{\text{s}}) 
} \right],
\end{equation}

Overall, the loss term $\mathcal{L}_{\text{seg}}$ in Eq.~\eqref{eq:loss1} applies an $L_1$ loss to penalize the absolute offset of anchor centers, enabling precise localization of event boundaries, while $\mathcal{L}_{\text{span}}$ aims to maximize the overlap between predicted spans and the ground truth, as suggested in~\cite{carion2020end}.
By minimizing the above anchor loss, the event localizer is trained to produce temporal anchors aligned with the ground-truth video segments. Later in the experiments, we will verify the effectiveness of this stage, allowing improved captioning performances.

\begin{algorithm}[t]
\caption{Temporal-Aware Video Event Captioning}
\label{alg:stage2}
\textbf{Model:} \\
Event Localizer $\theta_{\mathcal{E}}$, Time Embedding $\theta_{\mathcal{T}}$, LLM  $\mathcal{W}_{\text{LoRA}}$ \\
\textbf{Input:} \makebox[0pt][l]{Set of data $\mathcal{S}_v = \{ v^j \}^{| \mathcal{S}_v |}_{j=1}$}
\begin{algorithmic}[1]

    \STATE \textcolor{blue}{\textit{/* Learning to Predict Temporal Anchors */} }
    \FOR{each sample $v^j \in \mathcal{S}_v$}
        \STATE Video feature $h^{j}_v \leftarrow $ LLM(Visual\_Encoder($v^j$)) 
        \STATE Temporal Anchors $\hat{Y}^j \leftarrow \theta_{\mathcal{E}}(h^{j}_v)$
        \STATE Update $\theta_{\mathcal{E}}$ and $\theta_{\mathcal{T}}$ by $\mathcal{L}_{\text{anchor}}(Y, \hat{Y})$ 
    \ENDFOR

    
    \STATE \textcolor{blue}{\textit{/* Temporal-Aware Video Event Captioning */} }
    \FOR{each sample $v^j \in \mathcal{S}_v$}
        \STATE Video feature $h^{j}_v \leftarrow $ LLM(Visual\_Encoder($v^j$)) 
        \STATE Temporal Anchors $\hat{Y}^j \leftarrow \theta_{\mathcal{E}}(h^{j}_v)$
        \STATE $\tilde{Y}^j \leftarrow \text{LLM}(v^j, \hat{Y}^j)$
        \STATE  Update $\theta_{\mathcal{E}}$, $\theta_{\mathcal{T}}$, and $\mathcal{W}_{\text{LoRA}}$ by $\mathcal{L}_{\text{caption}}(C_i, \hat{C}_i)$
        \STATE Update $\theta_{\mathcal{E}}$ and $\theta_{\mathcal{T}}$ by  $\mathcal{L}_{\text{denoise}}(Y, \tilde{Y})$     
    
    \ENDFOR

\end{algorithmic}
\end{algorithm}

\subsubsection{Temporal-Aware Video Event Captioning}
\label{train:stage2}


As illustrated in Fig.~\ref{fig:main}, the temporal anchors produced by the event localizer are utilized as the input tokens to the LLM for dense captioning, via introducing a time embedding layer $\theta_{\mathcal{T}}$. In practice (for training and inference), such temporal anchors might contain noisy temporal information, and thus we have the LLM output the ground truth temporal tokens for guiding the learning of both $\theta_{\mathcal{E}}$ and $\theta_{\mathcal{T}}$. 

The above training scheme can be viewed as a \textit{denoising} process, encouraging the LLM to predict precise start/end time (and later the caption for the associated segment). Therefore, we extend~\cref{eq:loss1} and propose a denoising loss $\mathcal{L}_{\text{denoise}}$ as:
\begin{align} \label{eq:re_anchor_loss}
    \mathcal{L}_{\text{denoise}}(\tilde{Y},Y;\theta_{\mathcal{E}},\theta_{\mathcal{T}}) = \lambda_{\text{\scriptsize 1}} \mathcal{L}_{\text{\scriptsize seg}} + \lambda_{\text{\scriptsize 2}} \mathcal{L}_{\text{\scriptsize span}},
\end{align}
where $\tilde{Y}=(\tilde{c_i}, \tilde{d_i})$ denotes the event timestamps predicted from the hidden states of the LLM with a regression head.

On the other hand, to ensure the LLM's captioning ability under the presence of both video frames and temporal anchor, we train a LoRA~\cite{hu2021lora} module with a standard captioning loss $\mathcal{L}_{\text{caption}}$, which is defined as:

\begin{equation}\label{eq:caption_loss}
    \mathcal{L}_{\text{caption}} = \sum^{N}_{i=1} CE(C_i, \hat{C}_i; C_{<i}, \hat{Y}_{\leq i}, v),
\end{equation}



\noindent where the caption loss is a cross-entropy (CE) loss,  $C_i$ is the ground-truth event captions, $\hat{C_i}$ is the predicted event captions, $\hat{y}_{\leq i}$ is the utilized temporal anchors, $N$ is the ground-truth event number in the video $v$.

With the above anchor and captioning losses, our introduced event localizer, time embedding, and LoRA can be jointly trained. Together with the fixed visual encoder and LLM, the integrated network can be viewed as a VideoLLM. The overall training process is outlined in Algorithm~\ref{alg:stage2}. It is worth noting that, since the caption loss simply calculates a cross-entropy loss, simply using $\mathcal{L}_{\text{caption}}$ for training LoRA would not be sufficient to penalize the prediction error in prediction accuracy of time segments. This will also be verified later in our experiments. Comprehensive implementation details, including hyperparameters and architectural settings, are provided in Sec.~\ref{exp:implementation_details} and in the Appendix~\ref{supp:more_training_details} for reproducibility.

\subsection{Inference-time Event Coherent Sampling}
\label{method:inference}

With the above introduction, our \textit{TA-Prompting} is trained to exhibit the ability in exploiting temporal anchors for dense video captioning, with training objectives of identifying the anchors for each video segment and describing the corresponding video content. However, since the number of events within a video could be arbitrary during inference time, directly applying a fixed number ($M$) of temporal anchors would cause mismatched and overlapped event predictions, so that the resulting event captions lack global coherence throughout the whole video.

Motivated by the recent advancements in the decoding-time alignment of LLMs~\cite{liu2024decoding,shi2024decoding, wang2022self}, we propose an \textit{Event Coherent Sampling} strategy to enhance the temporal coherence during inference time. Given $M$ temporal anchors proposed by the event localizer, our sampling strategy uses a designed scoring function $\mathcal{S}$ to select one temporal anchor at a time for determining the output event-caption sequence. 

More specifically, a caption with a high confidence score during LLM's autoregressive generation indicates this predicted caption exhibits better consistency with the past prediction. Thus, we leverage this property as our selection score $\mathcal{S}_{\text{cs}}$ to allow the selection of coherent event captions during inference.
To further ensure the selected event captions are properly grounded to the given video, we employ the CLIP score~\cite{radford2021learning} $\mathcal{S}_{\text{as}}$ between the predicted caption and the corresponding video segment. 
Specifically, for each predicted segment, we compute the CLIP score~\cite{radford2021learning} between every frame image inside the segment and the corresponding event caption, and then take the mean of these scores to obtain $\mathcal{S}_{\text{as}}$.
As a result, selecting a temporal anchor with this scoring function enables \textit{TA-Prompting} to generate coherent event captions while regularizing \textit{TA-Prompting} to caption the corresponding video segment faithfully. 

The scoring function $\mathcal{S}$ is weighted sum of $\mathcal{S}_{\text{as}}$ and $\mathcal{S}_{\text{cs}}$, formulated as:
\begin{align} \label{eq:ecs}
\mathcal{S} &= \mathcal{S}_{\text{cs}} + \alpha \mathcal{S}_{\text{as}}, \text{ where}\\ 
\mathcal{S}_{\text{cs}} &= \exp(\frac{1}{|C_n|} \sum^{|C_n|}_{t=1} \log\left(p_{\mathcal{V}}(x_t | x_{<t}, C_{<n}, \hat{y}_{\leq n}, v)\right)), \\ 
\mathcal{S}_{\text{as}} &= \text{CLIP}(v_{\left[t^{s}_i : t^{e}_i\right]}, C_i)
= \frac{1}{|\mathcal{F}_i|}\sum_{f\in\mathcal{F}_i}
\mathrm{CLIP}\bigl(v_f,\,C_i\bigr).
\end{align}

Here, $\mathcal{F}_i$ denotes the set of frame indices within the interval $[t^s_i,t^e_i]$ of event $i$.
Note that $x_t$, $|C_n|$, $\hat{y}_{\leq n}$, and $v$ are defined in Eq.~\ref{eq:caption_loss}. And, $v_{[t^{s}_i : t^{e}_i]}$ and $C_{i}$ represent the video segment and the associated event caption. The score $\mathcal{S}_{cs}$ is a normalized conditional probability~\cite{wang2022self} that measures how well the VideoLLM $\mathcal{V}$ generates the $n^{\text{th}}$ event caption $C_n = \{x_1, \dots, x_{|C_n|}\}$, where each $x_t$ is a token in the caption. The generation is conditioned on the input video $v$, the previously generated captions $C_{<n}$, the previously predicted timestamps $\hat{y}_{<n}$, and the current timestamp $\hat{y}_n$.

By selecting the temporal anchor with the highest score according to $\mathcal{S}$, \textit{TA-Prompting} is able to avoid mismatched or overlapped event predictions, yielding better temporal dependency across predicted event captions.
We note that, for highly overlapping anchors from the event localizer, we perform clustering to obtain the representative event captions instead of directly applying all of the predicted anchors~\cite{kim2023self}. With this grouping strategy, the generation of repeated event captions could be effectively alleviated. More details related to implementation and ablation study about scoring component $\mathcal{S}$ are provided in the Appendix~\ref{supp:more_ecs}.
\section{Experiment}

\label{sec:experiment}

\subsection{Dataset and Experimental Setup}

\paragraph{Dataset} To evaluate our method's temporal grounding capabilities, we primarily conduct experiments on Dense Video Captioning (DVC) using ActivityNet Captions~\cite{krishna2017dense} and YouCook2~\cite{zhou2018towards} datasets. Since the evaluation of Dense Video Captioning primarily emphasizes the quality of the generated captions, we additionally assess the model's ability to localize events in videos by conducting experiments on the Event Localization task with ActivityNet. To further demonstrate our model's strength in temporal localization, we also evaluate it on the Moment Retrieval task, where the goal is to locate the timestamp corresponding to a given query within a video. We utilize the ActivityNet Captions~\cite{krishna2017dense} and Charades-STA~\cite{gao2017tall} datasets to assess the performance of VideoLLMs on this task. Furthermore, we adopt the task of TemporalQA to asses the event-based temporal understanding capability of VideoLLMs and perform the evaluation on ActivityNet-RTL~\cite{huang2024lita} dataset, which requires a model to generate a detailed response to a given question while accurately identifying the corresponding temporal segment.


\paragraph{Experimental Setup}
In Dense Video Captioning, we follow the experimental setups of VTimeLLM~\cite{huang2024vtimellm}, fine-tuning LITA~\cite{huang2024lita}, and VTG-LLM~\cite{guo2024vtg} on the ActivityNet Captions~\cite{krishna2017dense} and YouCook2~\cite{zhou2018towards} datasets for a fair comparison. For TimeChat~\cite{ren2024timechat}, we directly adopt the fine-tuned results reported in~\cite{guo2024vtg}. Meanwhile, we directly use VideoChat-7B~\cite{li2023videochat},  VideoChatGPT-7B~\cite{maaz2023video}, Momentor~\cite{qian2024momentor}, and TimeRefine~\cite{wang2024timerefine} for inference without further training. For Event Localization, we use the timestamps of all events generated by the Dense Video Captioning setting.

\begin{table*}[ht]
\centering
\vspace{-3mm}
\setlength{\tabcolsep}{2.1pt}
\renewcommand{\arraystretch}{1.5}
\begin{tabular}{l | ccc | ccc | cccc | cccc}
\toprule
& \multicolumn{6}{c|}{\textbf{Dense Video Captioning}} & \multicolumn{8}{c}{\textbf{Moment Retrieval}} \\
\cline{2-15}
\textbf{Models} & \multicolumn{3}{c|}{ActivityNet-Caption} & \multicolumn{3}{c|}{YouCook2} & \multicolumn{4}{c|}{ActivityNet-Caption} & \multicolumn{4}{c}{Charades-STA} \\
\addlinespace[-2pt]

& \footnotesize SODA\_c & \footnotesize CIDEr & \footnotesize METEOR & \footnotesize SODA\_c & \footnotesize CIDEr &\footnotesize  METEOR & \footnotesize R@0.3 & \footnotesize R@0.5 & \footnotesize R@0.7 & \footnotesize mIoU & \footnotesize R@0.3 & \footnotesize R@0.5 & \footnotesize R@0.7 & \footnotesize mIoU \\ 
\midrule
VideoChat-7B \cite{li2023videochat} & 0.9 & 2.2 & 0.9 & - & - & - & 8.8 & 3.7 & 1.5 & 7.2 & 9.0 & 3.3 & 1.3 & 6.5\\
VideoChatGPT-7B \cite{maaz2023video} & 1.9 & 5.8 & 2.1 & - & - & - & 26.4 & 13.6 & 6.1 & 18.9 & 20.0 & 7.7 & 1.7 & 13.7\\
\midrule
TimeChat-7B \cite{ren2024timechat} & 4.7 & 19.0 & 6.7 & 3.4 & 11.0 & - & 34.4 & 19.4 & 9.2 & 24.2 & 46.4 & 26.8 & 10.4 & 29.6 \\
Momentor-7B \cite{qian2024momentor} & 0.3 & 14.9 & 4.7 & - & - & - & 42.9 & 23.0 & 12.4 & 29.3 & 42.6 & 26.6 & 11.6 & 28.5 \\
LITA-13B \cite{huang2024lita} & 4.7 & 17.2 & 5.2 & 2.4 & 9.0 & 2.7 & 51.7 & 34.5 & 19.4 & 37.2 & 31.6 & 12.5 & 5.6 & 20.8 \\
VTG-LLM-7B \cite{guo2024vtg} & 5.1 & 20.7 & 5.9 & 3.6 & 13.4 & - & - & 8.3 & 3.7 & - & - & 33.8 & 15.7 & - \\
VTimeLLM-7B \cite{huang2024vtimellm} & 5.8 & 27.6 & 6.8 & 2.7 & 8.6 & 2.7 & 44.0 & 27.8 & 14.3 & 30.4 & 51.0 & 27.5 & 11.4 & 31.2 \\
TimeRefine-7B~\cite{wang2024timerefine} & \textbf{6.1} & 28.6 & \textbf{7.0} & - & - & - & 48.0 & 33.6 & 17.5 & 34.0 & \textbf{56.9} & \textbf{38.6} & 16.4 & \textbf{36.2} \\
\midrule
\textbf{TA-Prompting-7B (Ours)} & \textbf{6.1} & \textbf{29.2} & \textbf{7.0} & \textbf{4.3} & \textbf{15.6} & \textbf{4.2} & \textbf{52.6} & \textbf{34.9} & \textbf{20.5} & \textbf{37.5} & 48.8 & \underline{34.1} & \textbf{16.8} & \underline{32.3}  \\
\bottomrule
\end{tabular}
\caption{Qualitative evaluation of VideoLLMs on the tasks of Dense Video Captioning and Moment Retrieval. \textbf{Bold} and \underline{underline} indicate the best and second best results, respectively.}
\vspace{-3mm}
\label{tab:main}
\end{table*}

To apply \textit{TA-Prompting} to tackle the Moment Retrieval task, we prompt \textit{TA-Prompting} sequentially with each of the \(N\) temporal anchors identified by the event localizer, resulting in \(N\) generated captions—one for each temporal anchor. We then evaluate each caption using the ECS scoring function (as defined in Eq.~\ref{eq:ecs}), which calculates a score for each generated caption based on its alignment with the query. Among the \(N\) captions, we select the one with the highest ECS score. The temporal anchor associated with this highest-scoring caption is then used to define the predicted video segment, providing the temporal boundaries required for moment retrieval. As comparing performance on ActivityNet Captions~\cite{caba2015activitynet}, we ensure that all models, except VideoChat~\cite{li2023videochat}, VideoChatGPT~\cite{maaz2023video}, and Momentor~\cite{qian2024momentor}, are trained on this dataset. For Charades-STA~\cite{gao2017tall}, all models are evaluated in a zero-shot setting to assess their cross-domain generalization capabilities. For TemporalQA, we train our Temporal-Aware Video Event Captioning framework on the ActivityNet-RTL dataset~\cite{huang2024lita} and then adopt the inference pipeline used in Moment Retrieval.


\subsection{Evaluation Metrics}
For Dense Video Captioning, to evaluate the quality of generated captions, we calculate their alignment with ground truth captions at various Intersection over Union (IoU) thresholds ($iou = { 0.3, 0.5, 0.7, 0.9 }$). We utilize CIDEr~\cite{vedantam2015cider} and METEOR~\cite{banerjee2005meteor} as the evaluation metrics and report their averages over these IoU thresholds to provide a comprehensive performance assessment. Moreover, we also employ SODA\_c~\cite{fujita2020soda}, which evaluates both IoU and caption quality while preserving temporal order, to show the coherence of generated event captions. For Event Localization, we report Recall, Precision, and F1 scores. Recall is derived by averaging Recall@1 across different IoU thresholds ($iou={0.3,0.5,0.7,0.9}$), and Precision is also computed with the same set of IoU thresholds. The F1 score is then obtained as the harmonic mean of Recall and Precision. For Moment Retrieval, we Compute IoUs between predicted timestamps and ground truth. We report mean IoU (mIoU) and Recall@1 at IoU thresholds of $iou = \{ 0.3, 0.5, 0.7 \}$, which is denoted as $R@iou$. Following the evaluation protocol in LITA~\cite{huang2024lita} for Video Temporal Reasoning, we assess the quality of predicted intervals by reporting mean IoU (mIoU) and Precision@1 at an IoU threshold of 0.5, which is expressed as $P@0.5$. Furthermore, we employ GPT-4 to assign a score that evaluates the relevance of the predicted caption to the corresponding ground truth annotation.

\begin{table*}[ht]
\centering
\begin{minipage}[t]{0.48\linewidth} 
\centering
\vspace{-3mm}
\setlength{\tabcolsep}{1mm}
\begin{tabular}{c|ccc}
    \toprule
    Method & Recall & Precision & F1 \\
    \midrule
    LITA-13B~\cite{huang2024lita}   & 37.0 & 36.6 & 36.8 \\
    TimeChat-7B~\cite{ren2024timechat}   & 23.8 & 29.4 & 26.3 \\
    VTG-LLM-7B~\cite{guo2024vtg}   & 26.7 & 28.6 & 27.9 \\
    VTimeLLM-7B~\cite{huang2024vtimellm}   & 34.5 & 43.7 & 39.8 \\
    \midrule
    \textbf{TA-Prompting-7B (Ours)}   & \textbf{37.9} & \textbf{43.9} & \textbf{40.7} \\
    \bottomrule
\end{tabular}
\caption{Event localization on ActivityNet Captions.}
\label{table:EventLoc}
\end{minipage}
\hfill
\begin{minipage}[t]{0.48\linewidth}
\centering
\vspace{-3mm}
\setlength{\tabcolsep}{1mm}
\begin{tabular}{l|ccc}
\toprule
Model & mIOU & P@0.5 & Score \\
\midrule
Video-LLaMA-v2-13B~\cite{cheng2024videollama} & - & - & 32.1 \\
Video-ChatGPT-13B~\cite{maaz2023video} & - & - & 28.8 \\
\midrule
VTimeLLM-7B~\cite{huang2024vtimellm} & 19.9 & 18.5 & 30.2 \\
LITA-7B~\cite{huang2024lita} & 24.1 & 21.2 & \textbf{44.0} \\
\midrule
\textbf{TA-Prompting-7B (Ours)} & \textbf{25.5} & \textbf{23.9} & \textbf{44.0} \\
\bottomrule
\end{tabular}
\caption{Results of TemporalQA on ActivityNet-RTL.}
\label{table:2}
\end{minipage}
\end{table*}
\begin{figure*}
    \centering
    \includegraphics[width=0.8\linewidth]{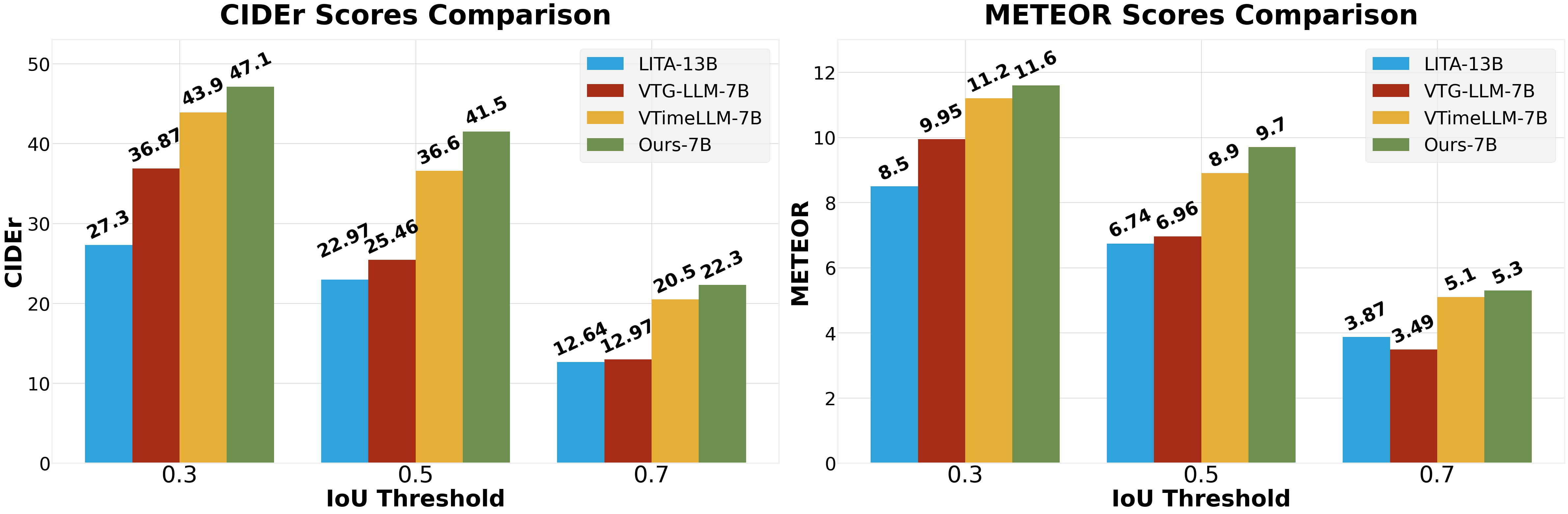}
    \caption{Evaluation of dense video captioning on the ActivityNet Captions datasets. The models to be compared with are \textcolor[HTML]{30a2da}{LITA} (blue), \textcolor[HTML]{a72c16}{VTG-LLM} (red), \textcolor[HTML]{e5ae38}{VTimeLLM} (yellow), and \textcolor[HTML]{6d904f}{\textit{TA-Prompting (Ours)}} (green). Note that the metrics of CIDEr and METEOR are calculated across different IoU thresholds of ${0.3, 0.5, 0.7}$.}
    \label{fig:ious}
    \vspace{-3mm}
\end{figure*}

\subsection{Implementation Details}
\label{exp:implementation_details}
The event localizer is composed of multiple event queries and cross-attention layers, with the number of queries set to exceed the maximum number of events in each dataset. Specifically, for ActivityNet~\cite{caba2015activitynet}, we use 10 queries and 3 cross-attention layers, each with a dimension of 512, while for YouCook2~\cite{zhou2018towards}, we use 50 queries and 2 cross-attention layers, each with a dimension of 512. For both datasets, the hyper-parameters of $\mathcal{L}_{anchor}$~\eqref{eq:loss1} are set as follows: $\lambda_1=10, \lambda_2=1$, following the previous approaches~\cite{lei2021detecting, moon2023query, jang2023knowing} since $\mathcal{L}_{seg}$ converges faster than $\mathcal{L}_{span}$. We adopt AdamW as our optimizer with a learning rate of 5e-5 and a weight decay of 1e-4, and employ StepLR as our learning rate scheduler. For stage A, we train ActivityNet for 50 epochs with batch size 128, and train YouCook2 for 100 epochs with batch size 128. All training process can be done on single RTX-3090 GPU.

We leverage Vicuna v1.5-7B~\cite{zheng2023judging} as our LLM backbone and train it using LoRA, following the LoRA settings from~\cite{huang2024vtimellm} and add LoRA to all linear modules except Event Localizer and Time Embedding. For video encoding, we use ViT-L/14~\cite{radford2021learning} to extract global features from uniformly sampled 100 frames per video, which are concatenated and pass through a linear layer to form the input video representation. This linear layer is pre-trained following the training process in~\cite{huang2024vtimellm} and then frozen along with ViT-L/14 during our proposed training stages. Our Time Embedding module follows the approach of~\cite{moon2023query, jang2023knowing}, leveraging sinusoidal positional encoding along with an MLP layer to project two-dimensional temporal anchors into LLM token embeddings. The encoded temporal anchors is then served as token semantic to the textual narratives. After passing LLM layers, we use the prediction head to generate a two-dimensional scalar from the LLM hidden feature, which is denoted as $\tilde{Y}$, and therefore we can use $\mathcal{L}_{\text{denoise}}$ to optimize our time embedding module. For reproducibility, comprehensive architectural details of our time embedding module are presented in the appendix.

Following the training process of Event Localizer, the hyper-parameters of $\mathcal{L}_{denoise}$ is set as follows: $\lambda_1=10, \lambda_2=1$. We use AdamW optimizer and cosine learning rate scheduler with learning rate set to 1e-4. We train our model for 3 epochs with batch size 8 along with gradient accumulation set to 4. All training process can be done on single RTX-3090 GPU. During training, to avoid the gradients of $\mathcal{L}_{denoise}$, which is irrelevant to language modeling, update our LoRA modules, we conduct back-propagation twice. For the first back-propagation, we update all trainable parameters with $\mathcal{L}_{caption}$, while for the second one, we freeze LoRA parameters and use $\mathcal{L}_{denoise}$ update the the remaining trainable modules, as depicted in Algorithm~\ref{alg:stage2}. More details on training \textit{TA-Prompting} can be found in the appendix.

\begin{figure*}[t]
    \vspace{-4mm}
    \centering
    \includegraphics[width=0.8\linewidth]{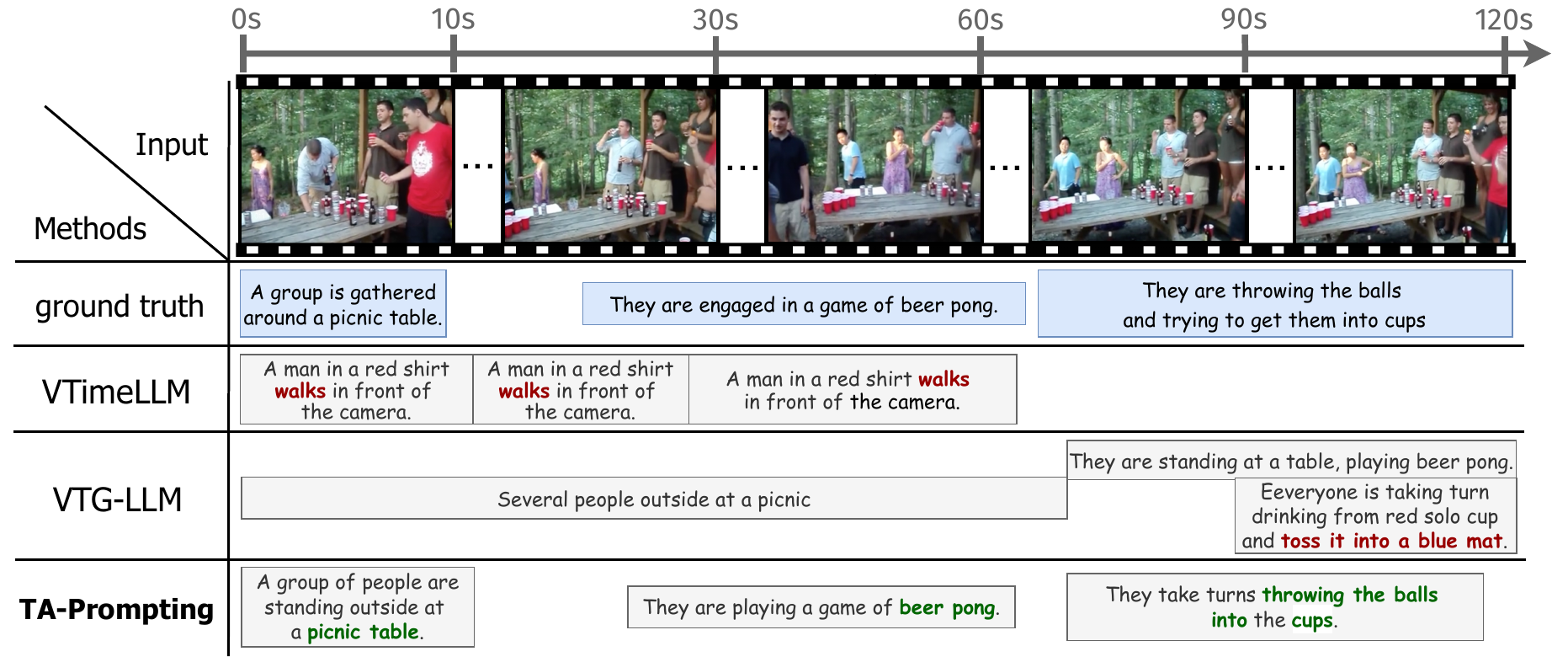}
    \caption{Qualitative evaluation and comparisons of predicted timestamps and the output captions. The width of each caption block represents the event duration, denoting the corresponding starting and ending time. Note that captions in green indicate correct alignment with the ground truth ones, while the captions in red indicate descriptions that are irrelevant to the visual content.}
    \label{fig:qualitative}
    \vspace{-3mm}
\end{figure*}

\subsection{Quantitative Evaluation}

We evaluate \textit{TA-Prompting} against recent temporal-understanding VideoLLMs across dense video captioning, event localization, moment retrieval, and temporal question answering.

\vspace{-3mm}
\paragraph{Dense Video Captioning.}
On ActivityNet Captions~\cite{krishna2017dense}, \textit{TA-Prompting} matches prior SOTA on SODA\_c and METEOR while improving CIDEr by +0.6. On YouCook2, it yields +0.7 SODA\_c, +2.2 CIDEr, and +1.5 METEOR. SODA\_c directly measures sequence-level coherence; moreover, CIDEr/METEOR at higher IoU thresholds indirectly reward temporally aligned, context-consistent captions. Consistently stronger results at higher IoUs (Fig.~\ref{fig:ious}) indicate superior modeling of coherent event sequences.

\vspace{-3mm}
\paragraph{Event Localization.}
Relative to the previous SOTA, \textit{TA-Prompting} improves Recall by +3.4, Precision by +0.2, and F1 by +0.9, showing that temporal anchors help detect a more complete set of events while refining each event’s time span.

\vspace{-3mm}
\paragraph{Moment Retrieval.}
Across ActivityNet Captions~\cite{krishna2017dense} and Charades-STA, \textit{TA-Prompting} outperforms other VideoLLMs, reflecting accurate timestamp grounding and strong transfer to related retrieval tasks. Its gains on Charades-STA further suggest promising zero-shot generalization.

\vspace{-3mm}
\paragraph{Temporal Question Answering.}
On ActivityNet-RTL~\cite{huang2024lita}, \textit{TA-Prompting} increases mIoU and Precision@0.5 while maintaining a comparable overall “score,” indicating better localization of the question-relevant interval together with detailed answer descriptions.

\subsection{Qualitative Evaluation}
In Figure~\ref{fig:qualitative}, we compare \textit{TA-Prompting} with VTimeLLM~\cite{huang2024vtimellm} and VTG-LLM~\cite{guo2024vtg} on ActivityNet Captions~\cite{krishna2017dense} using each model's official checkpoint and default settings. The video depicts a group playing beer pong. VTimeLLM repeatedly predicts~\textit{a man in a red shirt walks in front of the camera}, though he actually stands beside the table, and misses key actions like~\textit{playing beer pong} and~\textit{throwing the balls}. VTG-LLM also errs, captioning people as~\textit{tossing a red cup into a blue mat}, which conflicts with the segment. In contrast, \textit{TA-Prompting} accurately captures all major activities with faithful captions and precise temporal alignment.

\subsection{Ablation Study}

As shown in Table~\ref{tab:ablation}, we demonstrate the effectiveness of our primary components (i.e., Temporal-Aware Video Event Captioning and Event Coherent Sampling (ECS)), using the ActivityNet Captions dataset~\cite{krishna2017dense}. To assess their individual contributions, we incrementally remove each component from our whole version of \textit{TA-Prompting}.

Specifically, \textit{Ours w/o ECS} means that during inference, the model randomly selects temporal anchors to generate event captions, losing the coherence provided by ECS. The results indicate a noticeable decline across all evaluation metrics: SODA\_c decreases by 0.2, CIDEr by 2.2, and METEOR by 0.1. This underscores ECS's critical role in maintaining coherence across event captions. 

In addition, \textit{Ours w/o ECS and TA} further impacts performance by replacing temporal anchors with textual descriptions to represent timestamps. This leads to additional drops: SODA\_c decreases by 0.2, CIDEr by 0.3, and METEOR by 0.2, highlighting the importance of temporal anchors in providing high-quality temporal grounding. The results in Table~\ref{tab:ablation} successfully verify the effectiveness of our proposed ECS and temporal anchors.

Table~\ref{table:ecs_alpha} further varies the weight $\alpha$ in the ECS scoring function $\mathcal{S}=\mathcal{S}_{\text{cs}} + \alpha \mathcal{S}_{\text{as}}$ to assess the contribution of both LLM generation confidence and visual-textual alignment factors. The results show that TA-Prompting remains robust and consistently outperforms VTimeLLM~\cite{huang2024vtimellm}, achieving its best performance at $\alpha=1/5$.

\begin{table}
  \centering
  \begin{tabular}{@{}lccc@{}}
    \toprule
    \multirow{2}{*}{Method} & \multicolumn{3}{c}{ActivityNet-Caption} \\
    \cmidrule(lr){2-4}
    & SODA\_c & CIDEr & METEOR \\
    \midrule
    Ours (full) & \textbf{6.1} & \textbf{29.2} & \textbf{7.0} \\
    \hspace{2em} w/o ECS & 5.9 & 27.0 & 6.9 \\
    \hspace{2em} w/o ECS \& TA & 5.7 & 26.7 & 6.7 \\
    \bottomrule
  \end{tabular}
  \caption{Ablation study of our method. Note that \textbf{ECS} denotes \textbf{E}vent \textbf{C}oherent \textbf{S}ampling, while \textbf{TA} stands for \textbf{T}emporal \textbf{A}nchor.}
  \label{tab:ablation}
\end{table}

\begin{table}[ht]
  \centering
  \resizebox{0.8\columnwidth}{!}{%
  \begin{tabular}{@{}cccc@{}}
    \hline
     $\mathcal{S}_{\text{cs}} + \alpha \mathcal{S}_{\text{as}}$  & SODA\_c$\uparrow$ & CIDEr$\uparrow$ & METEOR$\uparrow$ \\ \hline
    $\alpha = 1$ & \textbf{6.1} & 28.7 & 6.8 \\
    $\alpha = 1/3$ & \textbf{6.1} & 29.1 & 6.9 \\
    $\alpha = 1/5$  & \textbf{6.1} & \textbf{29.2} & \textbf{7.0} \\
    \hline
  \end{tabular}%
    }
  \caption{Performance with different weight combinations in the ECS scoring function on ActivityNet-Caption benchmark.}
  \label{table:ecs_alpha}
  \vspace{-2mm}
\end{table}
\section{Conclusion}
\label{sec:conclusion}

In this paper, we proposed a novel VideoLLM framework of \textit{TA-Prompting}. With the introduced module of event localizer, \textit{TA-Prompting} is able to perform precise event localization while generating temporal-grounded dense captions. Our event localizer is designed to capture event boundaries by identifying temporal anchors, which steer the LLM to perform temporal-aware event captioning. With the goal of tackling arbitrary event numbers during inference, an event coherent sampling strategy is further proposed to determine the output event sequence while enforcing temporal coherence across events and semantic grounding to the corresponding video segment. We conducted extensive quantitative and qualitative evaluations on \textit{TA-Prompting}, 
verifying its satisfactory performance on both dense video captioning and moment retrieval tasks over several benchmark datasets.
\section{Acknowledgments}
This work is supported in part by the National Science and Technology Council via grant NSTC 114-2221-E-002-056-MY2 and NSTC 114-2640-E-002-006. We also thank the National Center for High-performance Computing (NCHC) for providing computational and storage resources.
{
    \small
    \bibliographystyle{ieeenat_fullname}
    \bibliography{main}
}
\clearpage
\setcounter{page}{1}
\maketitlesupplementary

\section{Detailed Evaluation Explanation} 
We mainly compare our model with VideoChat~\cite{li2023videochat}, VideoChatGPT~\cite{maaz2023video}, TimeChat~\cite{ren2024timechat}, Momentor~\cite{qian2024momentor}, LITA~\cite{huang2024lita}, VTimeLLM~\cite{huang2024vtimellm}, and VTG-LLM~\cite{guo2024vtg}. For VideoChat and VideoChatGPT, we directly use the scores reported in \cite{huang2024vtimellm}. For TimeChat, we leveraged the official checkpoint to evaluate its performance on both Moment Retrieval tasks. Momentor’s metric scores are also directly reported from their paper, where the model was evaluated in a zero-shot setting. For LITA, we generated results across all four datasets. Since ActivityNet-Caption was one of the domains LITA was originally trained on, we used its official checkpoint to conduct experiments on ActivityNet-Caption~\cite{krishna2017dense} for both DVC and MR tasks. For YouCook2, we fine-tuned LITA using LoRA to adapt it to the domain. Similarly, since VTimeLLM was not pretrained on the YouCook2 dataset, we also finetuned it using LoRA to ensure better alignment with the dataset characteristics. For other models not explicitly specified above, we report their metric scores directly as provided in their respective papers.

\section{Additional Details on Model Architecture and Training}
\label{supp:more_training_details}

\subsection{Details of Time Embedding Module}

We follow previous works~\cite{lei2021detecting, moon2023query, jang2023knowing}, applying sinusoidal positional encoding to map each 2-dimensional temporal anchor into a 4096-dimensional representation, matching the hidden size of Vicuna v1.5~\cite{zheng2023judging}. A two-layer MLP is then used to reduce the dimension to 512 and subsequently project it back to 4096, aligning with the LLM’s feature space. For reconstructing the event timestamp $\tilde{Y}$ from the LLM output, we reuse this same two-layer MLP, followed by a linear layer with output dimension 2, to predict the center and duration $(c_i, d_i)$ of each event.

\subsection{Temporal-Aware Video Event Captioning learning template}
\label{supp:tav_template}

For each event, we use the following format for captioning: ``$\texttt{<sep>}$$~\textcolor{red}{<s>}$$~\textcolor{blue}{<s>}$$~\text{`i-th event caption'}$.". The special token $\texttt{<sep>}$ marks the start of each event generation. For simplicity, we use ``*", a string not utilized during training, to represent $\texttt{<sep>}$ in our training template. The first $\texttt{<s>}$ token indicates the position to be replaced by the encoded temporal anchor, while the second $\texttt{<s>}$ marks the beginning of each event caption. Through this training template, \textit{TA-Prompting} effectively utilizes temporal anchors to generate accurate and contextually relevant event captions.

During self-attention within the LLM, we adjust the attention maps of the tokens corresponding to the temporal anchors such that they only attend to the video tokens. This design ensures that the temporal anchor focuses solely on the video context, as its primary function is to denoise and extract the relevant event timestamps from the video. By restricting its attention scope, TA-Prompting effectively refines the localization of events, ensuring that each generated caption is temporally aligned with the corresponding video segment.

\subsection{TA-Prompting's pre-training process}

Before fine-tuning \textit{TA-Prompting} on task specific objectives such as Dense Video Captioning and TemporalQA, we first pre-trained our LLM backbone (Vicuna v1.5-7B) with LoRA using Boundary Perception training data from VTimeLLM~\cite{huang2024vtimellm}, which is processed from InternVid~\cite{wang2023internvid}-10M-FLT. This pre-training step equips the LLM with an initial temporal awareness capability. To achieve this, we reformat the annotations to Temporal-Aware Video Event Captioning learning template, as depicted in Sec~\ref{supp:tav_template}, and train the LLM with LoRA for two epochs following this procedure: In the first epoch, we train the model using only textual supervision, leaving the first \texttt{<s>} tokens unchanged rather than replacing them with encoded temporal anchors. This allows the LLM to first learn the structure and linguistic patterns of event descriptions. In the second epoch, we apply the full training procedure, incorporating encoded temporal anchors to align captions with their corresponding temporal contexts. The LoRA configuration and training hyper-parameters remain identical to those detailed in Sec~\ref{exp:implementation_details}.

\section{The effectiveness of Temporal-Aware Video Event Captioning training}
In our main paper, we compare \textit{TA-Prompting} against various VideoLLMs, most of which do not include event localizer modules. To verify the effectiveness of our joint optimization, we combine VTimeLLM~\cite{huang2024vtimellm} with event localizer from PDVC~\cite{wang2021end}. During inference, we iteratively sample a timestamp from the generated proposals as input to VTimeLLM, where the selection probability of each proposal is weighted by its proposal score generated by PDVC. This process continues, with the chosen proposal guiding VTimeLLM in generating event descriptions, until the LLM terminates generation or all proposals are exhausted. The result is presented in Table~\ref{table:pdvc_vtime}, which indicates that the combination of event localizer and VideoLLM doesn't necessarily lead to performance improvement.

\begin{table}[ht]
    \centering
    \setlength{\tabcolsep}{1mm}
    \begin{tabular}{c|ccc}
        \toprule
        Method & CIDEr & METEOR & SODA\_c \\
        \midrule
        VTimeLLM-7B~\cite{huang2024vtimellm}   & 20.0 & 6.0 & 4.8 \\
        \textbf{TA-Prompting-7B (Ours)}   & \textbf{29.2} & \textbf{7.0} & \textbf{6.1} \\
        \bottomrule
    \end{tabular}
    \caption{Comparison of TA-Prompting and the combination of VTimeLLM with an event localizer on the ActivityNet Captions.~\cite{krishna2017dense}}
    \label{table:pdvc_vtime}
\end{table}

\section{Details of Event Coherent Sampling}
\label{supp:more_ecs}

The Event Coherent Sampling algorithm, as detailed in Section~\ref{method:inference}, aims to select the optimal temporal anchor for event caption generation. However, this original iterative process can be time-consuming, as it involves repeatedly evaluating similar temporal anchors, often leading to overlapping and repetitive captions that describe the same events.

To mitigate this in practical scenarios, we arrange the temporal anchors chronologically and group similar temporal anchors into clusters, forming a set of temporal anchor clusters. Each cluster represents a set of similar events, allowing us to process these anchors simultaneously. By expanding the batch size to include all anchors within a cluster, we generate captions in parallel, reducing redundant evaluations and improving efficiency. After processing a cluster, we proceed to the next one sequentially. We extend the batch size to ensure that each temporal anchor in the current cluster interacts with all previously generated event captions.

Although the decoding strategy depicted above reduces redundant computation, it introduces new challenges, such as the exponentially growing batch size and the complexity of selecting the final answer from multiple candidates. To address these issues, we employ the scoring function $\mathcal{S}$ defined in Eq.~\ref{eq:ecs}. If the batch size exceeds the certain batch size threshold during the expansion, we apply $\mathcal{S}$ to filter out captions with lower scores, thereby maintaining the batch size within the specified limit. A high score of $\mathcal{S}$ indicates that the generated caption aligns with the visual content and maintains coherence across consecutive events. By applying the modified Event Coherent Sampling with the scoring function $\mathcal{S}$ during the inference stage, we are able to produce captions that are coherent across events and visually faithful, while avoiding excessive computational costs, repeated descriptions, and contextually inconsistent outputs. The algorithm~\ref{alg:ecs} outline the overall process of Event Coherent Sampling approach.

\begin{algorithm}[t]
\caption{Event Coherent Sampling}
\label{alg:ecs}
\textbf{Model:} Event Localizer $\theta_{\mathcal{E}}$ \\
\textbf{Input:} video $v$
\begin{algorithmic}[1]
    \STATE Extract Temporal Anchors $\hat{Y}$ from video $v$ using Event Localizer $\theta_{\mathcal{E}}$ \\ 
    \STATE Sort the Temporal Anchors in order and group similar temporal anchors $\hat{Y}$ to form a set \textbf{X} of temporal anchor clusters. \\
    output = $\{ \}$
    \WHILE{(Temporal Anchor cluster set \textbf{X} not $\emptyset$)}
        \STATE Generate an event caption $C_i$ conditioned on each temporal anchor $(\hat{c}_i, \hat{d}_i)$ in the extracted temporal anchor cluster from \textbf{X} using a pop-left operation.
        \STATE calculate $\mathcal{S}_i$ for each event caption and select index $j$ that has highest score $\mathcal{S}_j$
        \STATE $\hat{Y}\leftarrow \hat{Y} - (\hat{c}_j, \hat{d}_j)$
        \STATE $\text{output} \leftarrow \text{output} + (t_j^s, t_j^e, C_j)$
        \IF{EOS in output}
            \STATE \textbf{break}
        \ENDIF
    \ENDWHILE

    \STATE \textbf{Output:} Final sequence $\{ (t_i^s, t_i^e, C_i) \}_{i=1}^{N}$ containing coherent event captions.
\end{algorithmic}
\end{algorithm}

To evaluate our inference time, we compare \textit{TA-Prompting} using Event Coherent Sampling (ECS) with VTimeLLM~\cite{huang2024vtimellm} using standard beam search (BS). Standard beam search is chosen as the comparison baseline because both ECS and BS involve the process of expanding a single input caption into multiple output sequences during inference. The comparison results are presented in Table~\ref{tab:time}, with the beam size for beam search set to 3. For this experiment, we use a single A100 GPU for each model and Python's built-in time package to measure inference time. The results show that our ECS achieves better performance while maintaining a similar computation time compared to beam search.

\begin{table}
  \centering
  \resizebox{0.98\columnwidth}{!}{%
  \begin{tabular}{@{}lcccc@{}}
    \toprule
    \multirow{2}{*}{Method} & \multirow{2}{*}{sec. per sample} & \multicolumn{3}{c}{ActivityNet-Caption} \\
    \cmidrule(lr){3-5}
    & & SODA\_c & CIDEr & METEOR \\
    \midrule
    TA-Prompting with \textbf{ECS} & 7.08 & \textbf{6.1} & \textbf{29.2} & \textbf{7.0} \\
    VTimeLLM with \textbf{BS} & 6.98 & 5.8 & 27.6 & 6.8 \\
    \bottomrule
  \end{tabular}%
  }
  \caption{Comparison of inference time (second per sample) between standard beam search (BS) and proposed event coherence sampling (ECS) on ActivityNet Caption dataset~\cite{krishna2017dense}.}
  \label{tab:time}
  \vspace{-2mm}
\end{table}

\begin{figure*}
    \centering
    \includegraphics[width=1\linewidth]{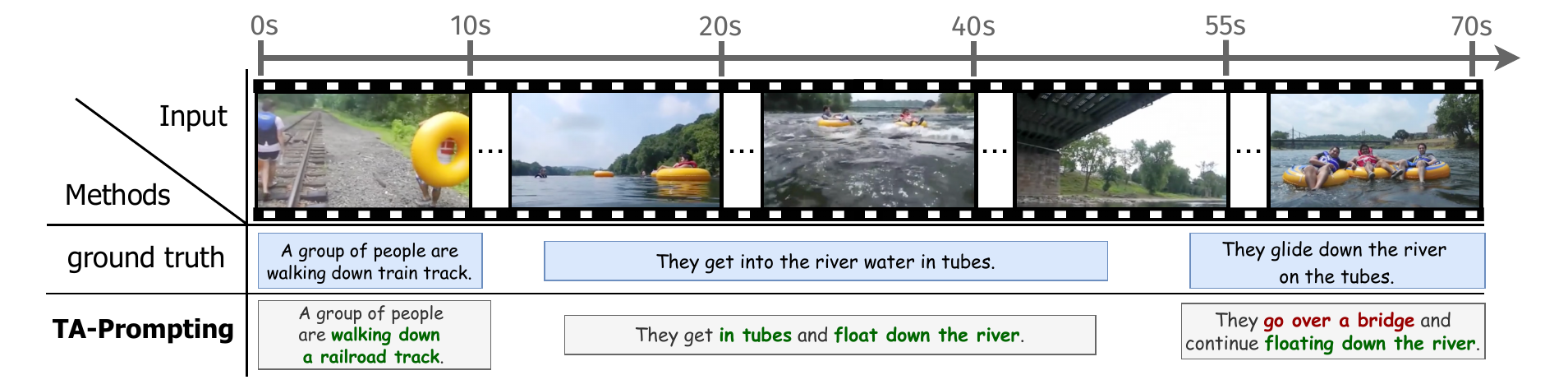}
    \caption{Visualization of a failure case with qualitative comparison of predicted timestamps and captions. The captions in green indicate correct alignment with the ground truth ones, while
the captions in red indicate descriptions that are irrelevant to the visual content.}
    \label{supp:hallucinate}
\end{figure*}

\section{Specifying the inputs to the event localizer}

Following LM4VisualEncoding~\cite{pang2023frozen}, we use the LLM’s output video features as visual inputs (refer to Figure~\ref{fig:main}) to the Event Localizer, demonstrating that LLMs can focus on more informative visual tokens. During inference, video features are processed by the LLM, and the resulting features are fed into the Event Localizer to generate temporal anchors. These anchors are then sequentially provided to the LLM for event caption generation. Since video features pass through the LLM only once, the inference efficiency of our model is comparable to that of general VideoLLMs.

\section{Choice of Timestamp Prediction During Inference}

In our implementation, we ultimately use the temporal anchors $\hat{Y}$ generated by the Event Localizer as the predicted timestamps during inference, rather than the denoised $\tilde{Y}$. This decision is supported by our experimental results, where we observe comparable performance using $\tilde{Y}$ (SODA\_c \textbf{6.1}, CIDEr \textbf{29.3}, METEOR 6.9) and temporal anchors $\hat{Y}$ (SODA\_c \textbf{6.1}, CIDEr 29.2, METEOR \textbf{7.0}). The similarity in performance can be attributed to the effectiveness of the Event Confidence Scorer (ECS), which helps filter out noisy temporal anchors during the decoding process. As a result, both options are reliable for our final timestamp prediction.

\section{Possibility of generalization to other datasets}
\begin{figure}
    \centering
    \includegraphics[width=1\linewidth]{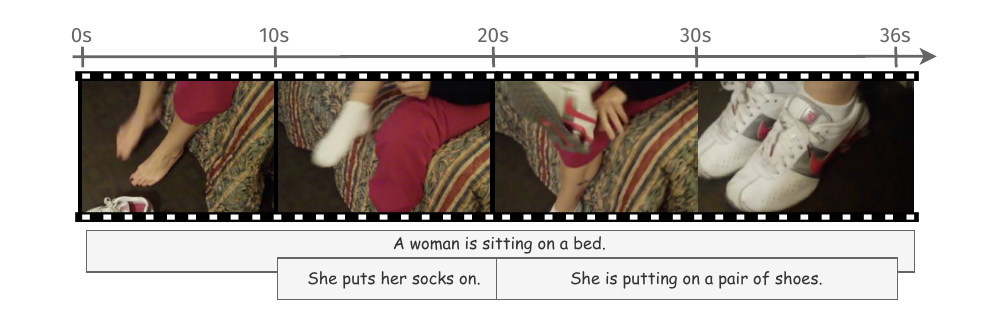}
    \caption{Visualization of dense captions on an out-of-domain YouTube video generated by TA-Prompting. Based on our observation, the video shows a woman sitting on a bed, sequentially putting on socks and then shoes. The visualization demonstrates that TA-Prompting accurately captures the key events occurring in the video.}
    \label{supp:random_video}
\end{figure}

To explore the generalization ability of TA-Promping beyond its training data (ActivityNet-Caption~\cite{krishna2017dense} and YouCook2~\cite{zhou2018towards}), we randomly selected a video from YouTube and present result in Figure~\ref{supp:random_video}. Although there is no ground truth in this sample, the visualization shows that TA-Prompting has the potential to generalize to out-of-domain tasks.

\section{Limitation and Failure Cases}

In our work, we represent video content by extracting CLIP features from uniformly sampled frames and concatenating them into a single visual feature sequence. These visual features could potentially be improved by leveraging recent VLLM methods developed for long video understanding. Another limitation of our approach is the risk of hallucinated captions, which may result from the choice of LLM backbone. As illustrated in Figure~\ref{supp:hallucinate}, \textit{TA-Prompting} can sometimes generate descriptions that are unrelated to the actual video content.

\section{More Quantitative Results}

\begin{figure*}
    \centering
    \includegraphics[width=1\linewidth]{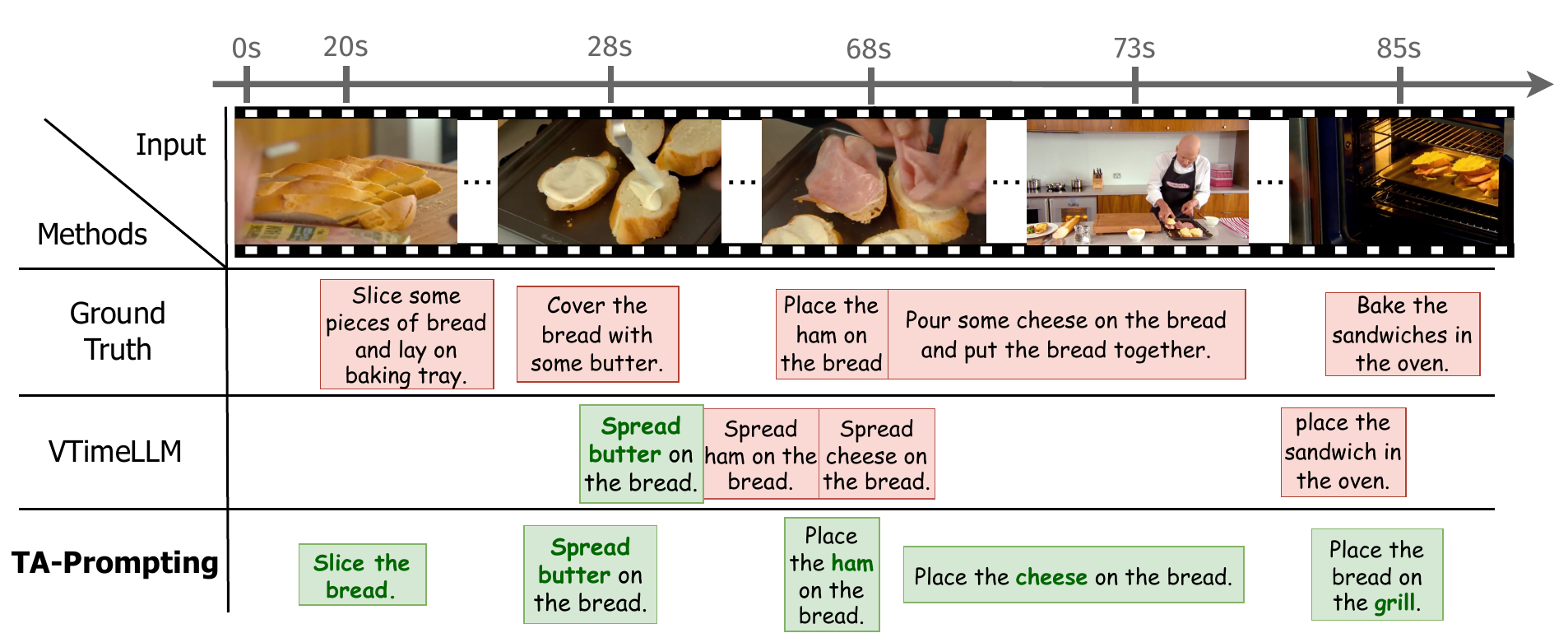}
    \caption{Qualitative results of the dense video captioning task on the Youcook2 dataset~\cite{zhou2018towards}. We compare TA-Prompting with VTimeLLM~\cite{huang2024vtimellm}. The width of each stripe indicates the predicted duration by each model.}
    \label{fig:enter-label}
\end{figure*}

\begin{figure*}
    \centering
    \includegraphics[width=1\linewidth]{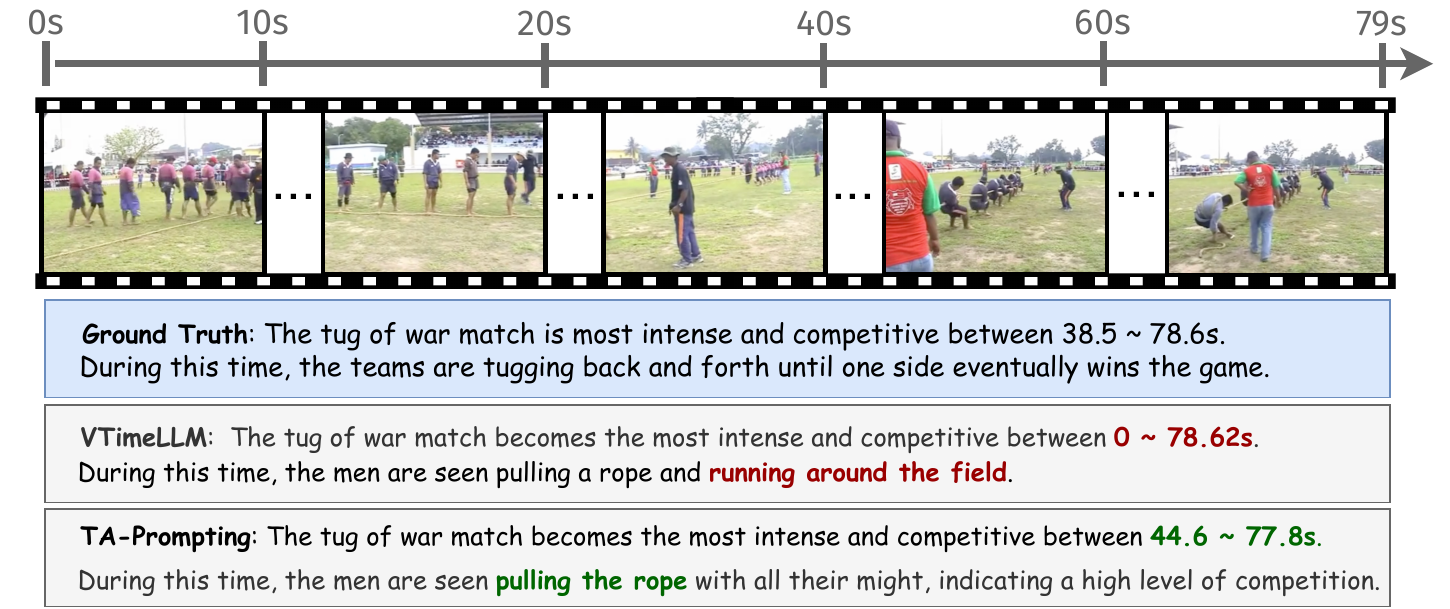}
    \caption{Qualitative results on ActivityNet-RTL~\cite{huang2024lita}. The captions in green indicate correct alignment with the ground truth ones, while the captions in red indicate descriptions that are irrelevant to the visual content.}
    \label{fig:anet_rtl}
\end{figure*}

\begin{figure*} 
    \centering
    \includegraphics[width=1\linewidth]{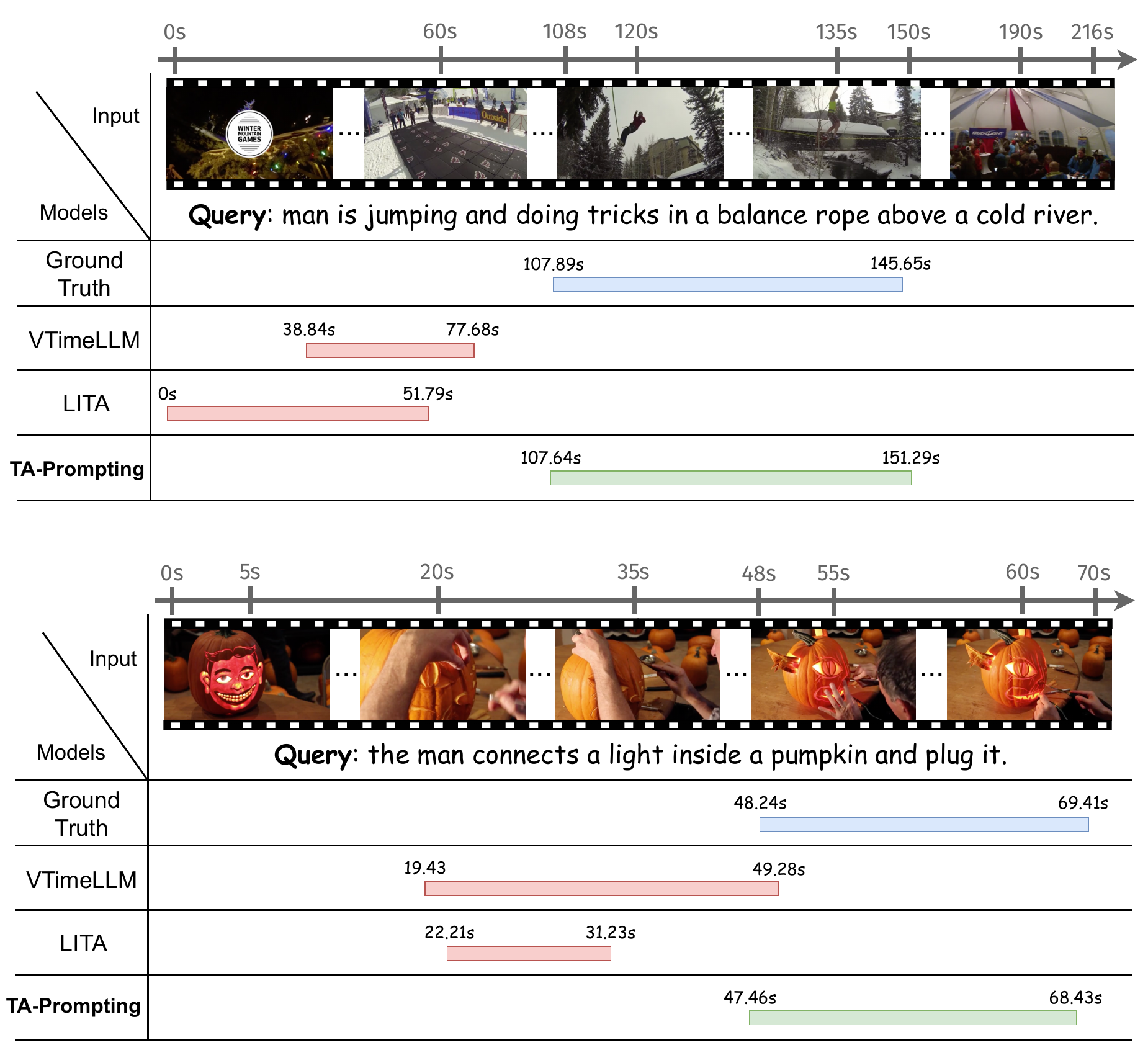}
    \caption{Qualitative results of the moment retrieval task on the ActivityNet-Caption dataset~\cite{caba2015activitynet}. We compare TA-Prompting with two previous state-of-the-art models: LITA~\cite{huang2024lita} and VTimeLLM~\cite{huang2024vtimellm}. The width of each stripe indicates the predicted duration by each model, with the start and end times labeled at the beginning and end of the stripe, respectively.}
    \label{fig:anetmr}
\end{figure*}

\begin{figure*} 
    \centering
    \includegraphics[width=1\linewidth]{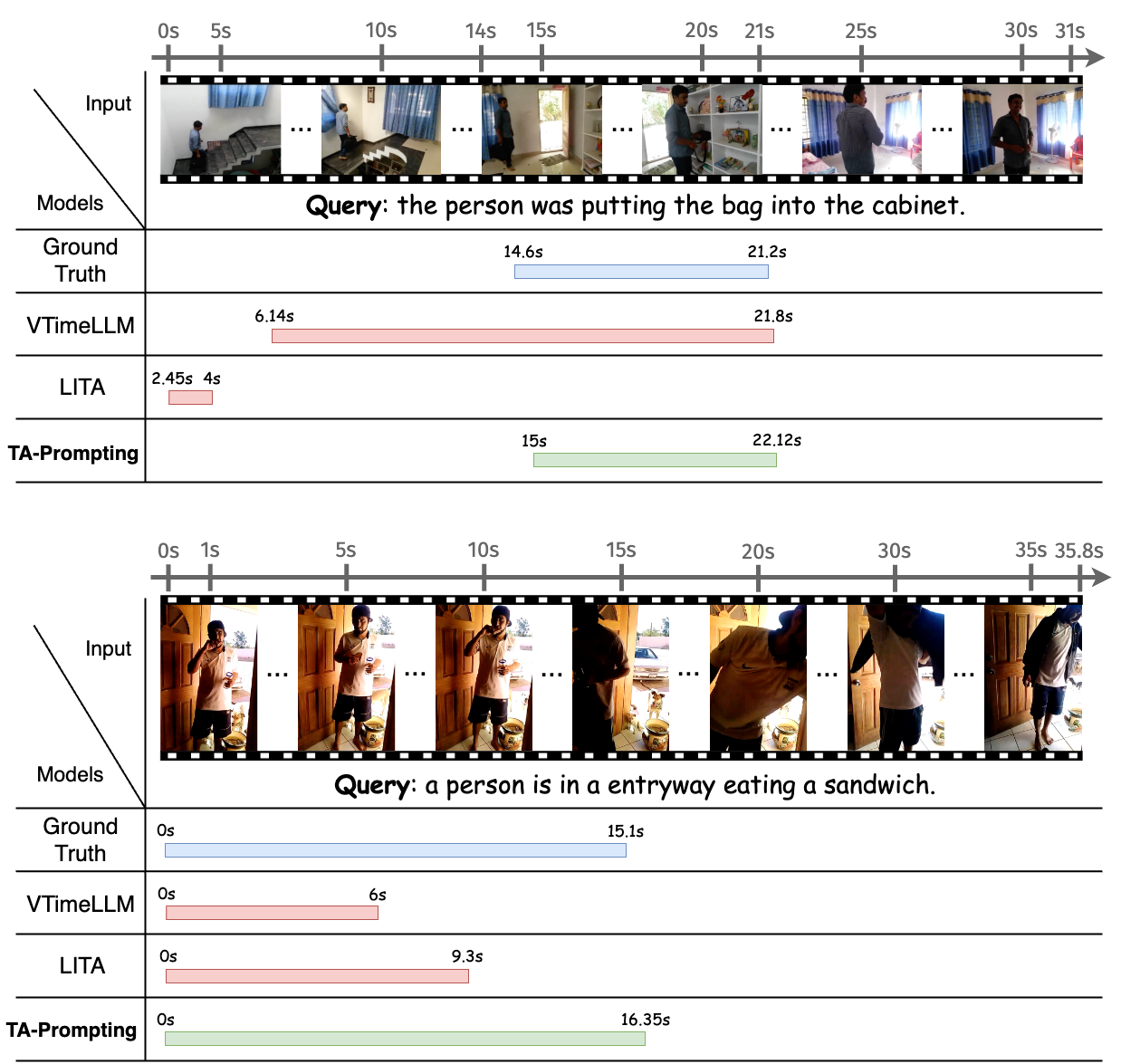}
    \caption{Comparison of TA-Prompting with two previous state-of-the-art models, LITA~\cite{huang2024lita} and VTimeLLM~\cite{huang2024vtimellm}, on the moment retrieval task using the Charades-STA dataset~\cite{gao2017tall}. Each stripe represents the predicted segment by the respective models, with the beginning and end points clearly marked to indicate the retrieved duration.}
    \label{fig:charades}
\end{figure*}

We present additional visualization results to provide a clearer comparison of our model's performance. For the dense video captioning task, we further show the visualization result on the YouCook2~\cite{zhou2018towards} dataset compared with VTimeLLM~\cite{huang2024vtimellm}, as shown in Figure~\ref{fig:enter-label}. For the moment retrieval task, we compare our results against VTimeLLM~\cite{huang2024vtimellm} and LITA~\cite{huang2024lita}, both of which have demonstrated significant performance among all the VideoLLM models used for our comparisons in Table~\ref{tab:main}. 
Figure~\ref{fig:anetmr}, illustrate the results on AcivityNet-Caption~\cite{caba2015activitynet}, while Figure~\ref{fig:charades} shows the results on Charades-STA~\cite{gao2017tall} dataset. Figure~\ref{fig:anet_rtl} shows the results of ActivityNet-RTL~\cite{huang2024lita}.

\end{document}